\documentclass{article}

\usepackage{microtype}
\usepackage{graphicx}
\usepackage{subfigure}
\usepackage{booktabs} 
\usepackage{hyperref}
\usepackage{color}

\usepackage{amssymb,dsfont,amsmath,amsthm,mathtools,natbib}
\usepackage[accepted]{aistats2019}

\DeclareMathOperator{\Tr}{Tr}

\def\R{\mathbb{R}}

\def\x{\mathbf{x}}

\def\L{\mathbf{L}}

\def\E{\mathbf{E}}

\def\m{\mathbf{m}}

\def\mX{\mathcal{X}}
\def\mA{\mathcal{A}}
\def\mD{\mathcal{D}}
\def\mE{\mathcal{E}}

\def\v{\mathbf{v}}

\usepackage{amsmath}

\theoremstyle{definition}
\newtheorem{defn}{Definition}
\newtheorem{theorem}{Theorem}

\newtheorem{lemma}{Lemma}

\begin{document}

\twocolumn[
\aistatstitle{Distance Measure Machines}

\aistatsauthor{ Alain Rakotomamonjy  \And  Abraham Traor\'e \And  Maxime B\'erar
\And R\'emi Flamary \And Nicolas Courty}

\aistatsaddress{ LITIS EA4108 \\Universit\'e de Rouen \\ and \\ Criteo AI Labs \\ Criteo Paris
 \And   LITIS EA4108 \\ Universit\'e de Rouen \And LITIS EA4108 \\
  Universit\'e de Rouen
 \And Lagrange UMR7293 \\ Universit\'e de \\Nice Sophia-Antipolis \And IRISA UMR6074 \\
 Universit\'e de \\ \\Bretagne-Sud
}
]

\begin{abstract}
  This paper presents a distance-based discriminative framework for
  learning with probability distributions. Instead of using kernel
  mean embeddings or generalized radial basis kernels, we introduce
  embeddings based on dissimilarity of distributions to some reference
  distributions denoted as templates. Our framework extends the theory
  of similarity of \citet{balcan2008theory} to the population
  distribution case and we show that, for some learning problems, some
  dissimilarity on distribution achieves low-error linear decision
  functions with high probability. Our key result is to prove that the
  theory also holds for empirical distributions. Algorithmically, the
  proposed approach consists in computing a mapping based on pairwise
  dissimilarity where learning a linear decision function is amenable.
  Our experimental results show that the Wasserstein distance
  embedding performs better than kernel mean embeddings and computing
  Wasserstein distance is far more tractable than estimating pairwise
  Kullback-Leibler divergence of empirical distributions.
 
\end{abstract}

\section{Introduction}

\begin{figure*}[t]
  \centering
  \includegraphics[width=0.36\linewidth]{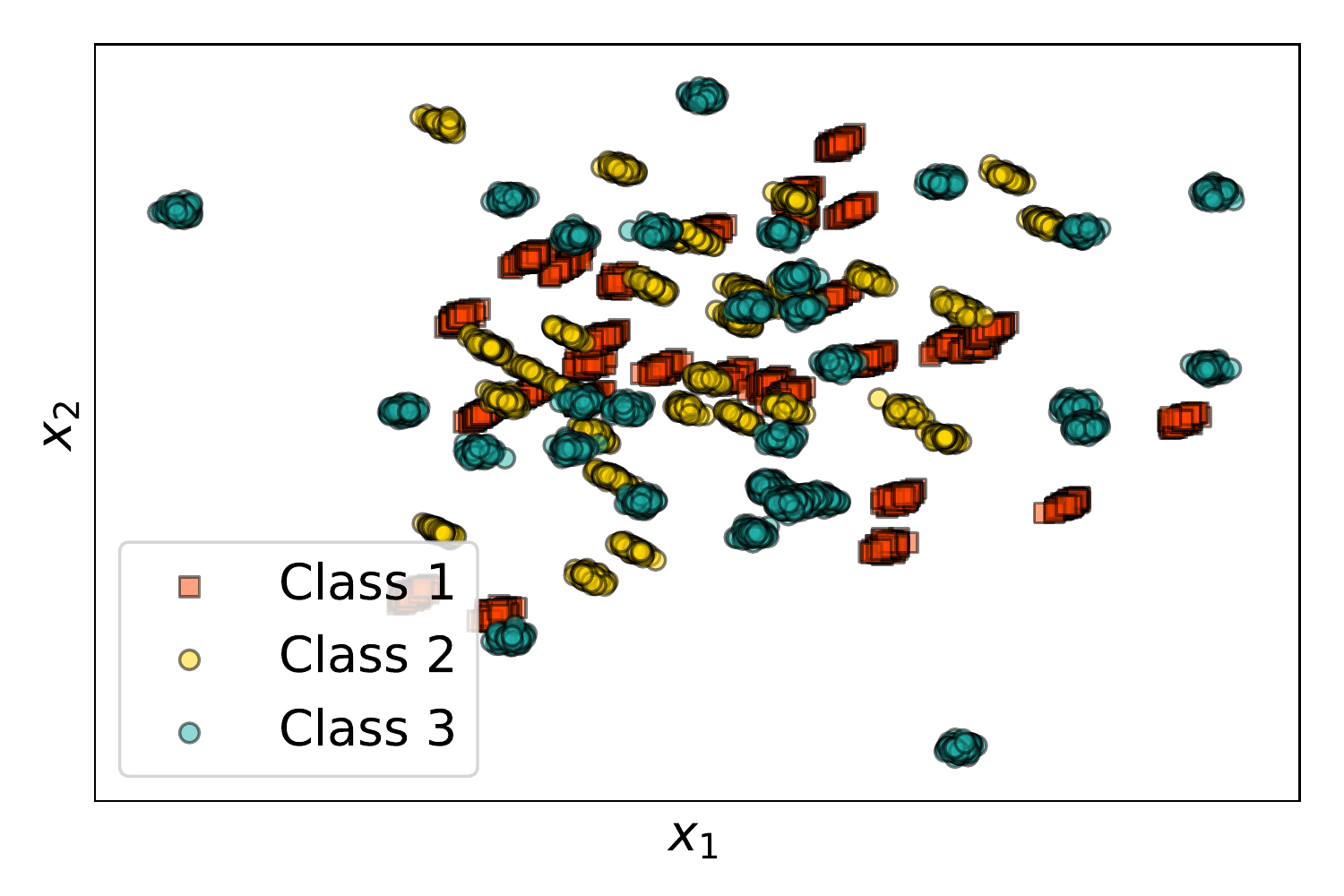}
  \includegraphics[width=0.63\linewidth]{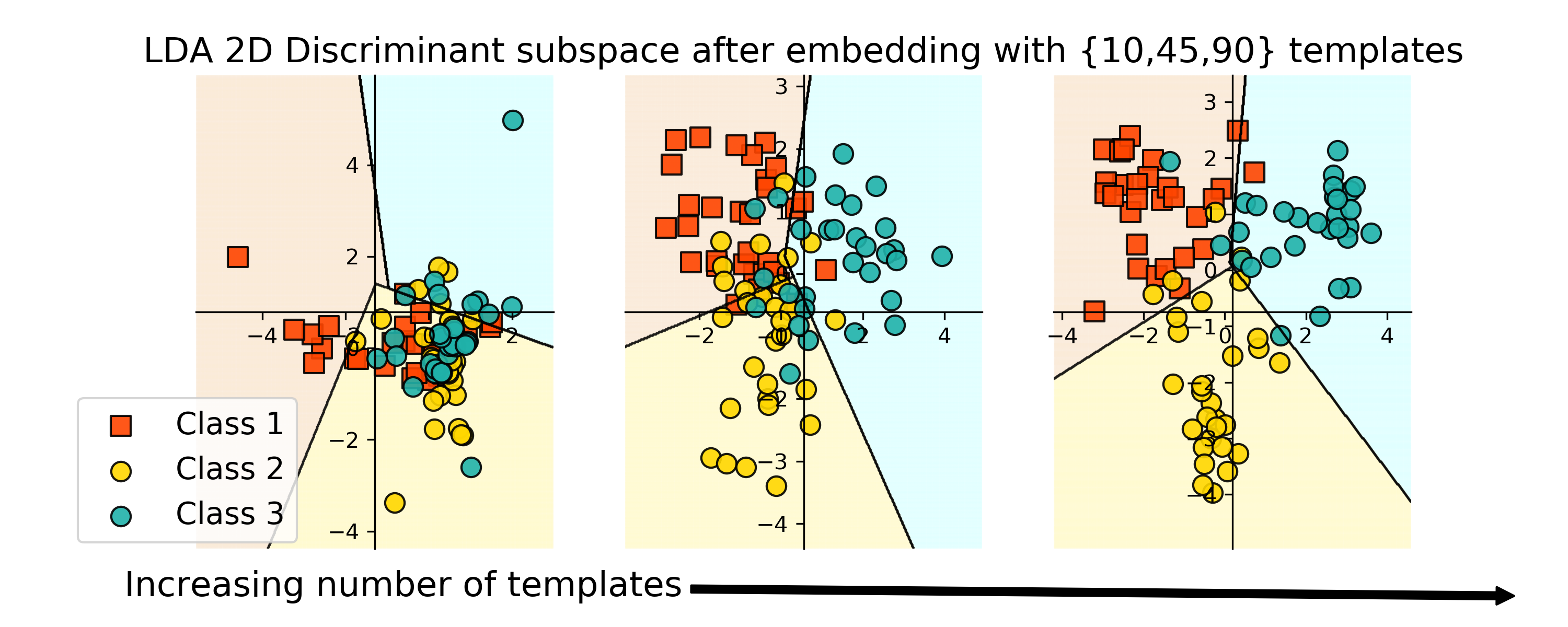}

  \caption{Illustrating the principle of the dissimilarity-based
    distribution embedding.  We want to discriminate empirical normal
    distributions in $\R^2$; their discriminative feature being the
    correlation between the two variables. An example of these normal
    distributions are given in the left panel. The proposed approach
    consists in computing a embedding based on the dissimilarity of
    all these empirical distributions (the blobs) to few of, them that
    serve as templates. Our theoretical results show that if we take
    enough templates and there is enough samples in each template then
    with high-probability, we can learn a linear separator that
    yields few errors.  This is illustrated in the 3 other panels in
    which we represent each of the original distribution as a point
    after projection in an discriminant 2D space of the embeddings.
    From left to right, the dissimilarity embedding respectively
    considers $10$, $45$ and $90$ templates and we can indeed
    visualize that using more templates improve separability.}
    \label{fig:intro}
\end{figure*}

Most discriminative machine learning algorithms have focused on learning
problems where inputs can be represented as feature vectors of
fixed dimensions. This is the case of popular algorithms like
support vector machines \citep{scholkopf2002learning} or random forest  \citep{breiman2001random}.
However, there exists several practical situations where it
makes more sense to consider input data as set of distributions
or empirical distributions instead of a larger collection of single vector. 
As an example, multiple instance learning \citep{dietterich1997solving} can be seen as 
learning of a bag of feature vectors and each bag can be interpreted
as samples from an underlying unknown distribution. 
Applications related to political sciences \citep{flaxman2015supported} or astrophysics \cite{ntampaka2015machine} 
have
also considered this learning from distribution point of view for
solving some specific machine learning problems. This paper also addresses
the problem of learning decision functions that discriminate distributions.

Traditional approaches for learning from distributions
is to consider reproducing kernel Hilbert spaces (RKHS) and associated kernels on
distributions. In this larger context, several kernels on distributions
have been proposed in the literature such as the probability product
kernel \citep{jebara2004probability}, the Battarachya kernel \citep{bhattacharyya1943measure} or the Hilbertian kernel on probability measures of  \citet{hein2005hilbertian}. 
Another elegant approach for kernel-based distribution learning has been proposedby \citet{muandet2012learning}. It consists in defining an explicit embedding
of a distribution as a mean embedding in a RKHS. Interestingly, if the 
kernel of the RKHS satisfies some mild conditions then all the information
about the distribution is preserved by this mean embedding. Then owing
to this RKHS embedding, all the machinery associated to kernel machines
can be deployed for learning from these (embedded) distributions.

By leveraging on the flurry of distances between distributions \citep{sriperumbudur2010non}, it is also possible to build
definite positive kernel by considering generalized radial basis function kernels of the form 
\begin{equation}
  \label{eq:genrbf}
  K(\mu,\mu^\prime) = e^{-\sigma d^2(\mu,\mu^\prime)},
\end{equation}
where $\mu$ and $\mu^\prime$ are two distributions, $\sigma>0$ a parameter
of the kernel and $d(\cdot,\cdot)$ a distance
between two distributions satisfying some appropriate properties so as to make
$K$ definite positive \citep{haasdonk2004learning}.

As we can see, most works in the literature address the question
of discriminating distributions by considering either implicit or
explicit kernel embeddings. 
However, is this really necessary?  Our observation is that there are many advantages of directly using distances or even dissimilarities between distributions for learning. 
It  would  avoid the need for two-stage approaches, computing the distance and then the kernel, as proposed by  \citet{poczos2013distribution} for distribution regression
or estimating the distribution and computing
the kernel as introduced by \citet{sutherland2012kernels}. Using kernels limits the choice of
distribution distances as the resulting kernel has to be definite positive.
For instance,  \citet{poczos2012nonparametric} used Reyni divergences for building generalized RBF kernel that turns out to be non-positive. For the same reason, the celebrated and widely used Kullback-Leibler divergence does not qualify for being used in a kernel.
This work aims at showing that learning from distributions with distances or even dissimilarity is indeed possible. Among all available distances on distributions, we focus our analysis on Wasserstein distances which come with several relevant properties,  that we will highlight later,  compared to other ones (\emph{e.g}  Kullback-Leibler divergence).

Our contributions, depicted graphically in Figure~\ref{fig:intro}, are the following : (a)  We show that by following the underlooked works of
 \citet{balcan2008theory}, learning to discriminate population distributions with dissimilarity functions comes at no expense. 
While this might be considered a straightforward extension, we are not aware of any work making this
connection. 
(b) Our key theoretical contribution is to show that Balcan's
framework also holds for empirical distributions if the used dissimilarity
function is endowed with nice convergence properties of the distance
of the empirical distribution to the true ones. 
(c) While these convergence bounds have already been exhibited for
distances such as the Wasserstein distance or MMD, we prove that
this is also the case for the Bures-Wasserstein metric.(d) Empirically, we illustrate the benefits of using this Wasserstein-based dissimilarity functions compared
to kernel  or MMD distances  in some simulated and real-world vision
problem, including 3D point cloud classification task.

 \section{Framework}
In this section, we introduce the global setting and present the
theory of learning with dissimilarity functions of \citet{balcan2008theory}.

\subsection{Setting}

Define $\mX$ as an non-empty subset of $\R^d$ and let $\mathbb{P}$ denotes
the set of all probability measures on a measurable space $(\mX, \mA)$,
where $\mA$ is  $\sigma$-algebra of subsets of $\mX$.  
Given a training set $\{\mu_i, y_i \}_{i=1}^n$, where $\mu_i \in \mathbb{P}$
and $y_i \in \{-1,1\}$, drawn \emph{i.i.d} from a probability distribution $P$
on $\mathbb{P} \times \{-1,1\}$, our objective is to learn a decision
function $h : \mathbb{P} \mapsto \{-1,1\}$ that predicts the most accurately
as possible the label associated to a novel measure $\mu$. In summary, our
goal is to learn to classify probability distributions from a supervised
setting. While we focus on a binary classification, the framework we consider
and analyze can be extended to multi-class classification.

\subsection{Dissimilarity function}

Most learning algorithms for distributions are based on reproducing kernel
Hilbert spaces and leverage on kernel value $k(\mu, \mu^\prime)$ between
two distributions where $k(\cdot,\cdot)$ is the kernel of a given RKHS.

We depart from this  approach and instead, we consider learning algorithms
that are built from pairwise dissimilarity measures between distributions. 
Subsequent definitions and theorems are recalled from \citet{balcan2008theory} and
adapted so as to suit our definition of bounded dissimilarity.

\begin{defn} A dissimilarity function over $\mathbb{P}$ is any 
pairwise function $\mD : \mathbb{P} \times \mathbb{P} \mapsto [0, M]$.
\end{defn}

While this definition emcompasses many functions, given two
probability distributions $\mu$ and $\mu^\prime$, we expect
$\mD(\mu,\mu^\prime)$ to be large when the two distributions
are ``dissimilar'' and to be equal to $0$ when they are similar. As such
any bounded distance over $\mathbb{P}$ fits into our notion
of dissimilarity, eventually after rescaling. Note that
unbounded distance which is clipped above $M$ also fits
this definition of dissimilarity.

Now, we introduce the definition
that characterizes dissimilarity function that allows one to learn a decision
function producing low error for a given learning task.

\begin{defn}\cite{balcan2008theory} \label{def:diss} A dissimilarity function $\mD$ is a
$(\epsilon,\gamma)$-good dissimilarity function for a learning problem
$\L$ if there exists a bounded weighting function $w$ over
$\mathbb{P}$, with $w(\mu) \in [0,1]$ for all $\mu \in \mathbb{P}$, such
that a least $1-\epsilon$ probability mass of distribution examples
$\mu$ satisfy :
$\E_{\mu^\prime \sim P} [w(\mu^\prime)\mD(\mu, \mu^\prime)| \ell(\mu) = \ell(\mu^\prime)] + \gamma \leq  \E_{\mu^\prime \sim P} \left[w(\mu^\prime)\mD(\mu, \mu^\prime)| \ell(\mu) \neq \ell(\mu^\prime) \right].$
The function $\ell(\mu)$ denotes the true labelling function that
maps $\mu$ to its labels $y$.
\end{defn}

In other words, this definition translates into: a dissimilarity function is ``good'' if with high-probability, the weighted average
of the dissimilarity of one distribution to those of the same label
is smaller with a margin $\gamma$ to the dissimilarity of distributions
from the other class.  

As stated in a theorem of \citet{balcan2008theory}, such a good
dissimilarity function can be used to define an explicit mapping 
of a distribution into a space. Interestingly, it can be shown that there exists
in that space a linear separator that produces low errors.
 
\begin{theorem} \cite{balcan2008theory} if $\mD$ is an $(\epsilon,\gamma)$-good dissimilarity
function, then if one draws a set $S$ from $\mathbb{P}$ containing
$n = (\frac{4M}{\gamma})^2 \log(\frac{2}{\delta})$ positive examples $S^+ = \{\nu_1, \cdots, \nu_n\}$ and 
$ n$ negative examples  $S^- = \{\zeta_1, \cdots, \zeta_n\}$, then with probability 
$1 - \delta$, the mapping $\rho_S : \mathbb{P} \mapsto \R^{2n}$
defined as  $\rho_S(\mu) = (\mD(\mu, \nu_1 ), \cdots, \mD(\mu, \nu_n),
\mD(\mu, \zeta_1 ), \cdots, \mD(\mu, \zeta_n ) )$ has the property
that the induced distribution $\rho_S(\mathbb{P})$ in $\R^{2n}$
has a separator of error at most $\epsilon + \delta$ at
margin at least $\gamma/4$.
\label{th:main}
\end{theorem}

The above described framework shows that under some mild conditions
on a dissimilarity function and if we consider
population distributions, then we can  benefit from the mapping $\rho_S$.  
However, in practice, we do have access only to empirical version
of these distributions. Our key theoretical contribution in Section \ref{sec:empirical} proves that if the number of distributions $n$ is large
enough and enough samples are obtained from each of these
distributions, then this framework is applicable with theoretical
guarantees to empirical distributions.

\section{Learning with empirical distributions}
\label{sec:empirical}
In what follows, we formally show under which conditions
an  $(\epsilon,\gamma)$-good dissimilarity function for some learning problems, applied to empirical distributions also produces
a mapping inducing low-error linear separator.

Suppose that we have at our disposal a dataset
composed of $\{\mu_i, y_i =1\}_{i=1}^{n}$ where each $\mu_i$ is a distribution.
However, each $\mu_i$ is not observed directly but instead we observe
it empirical version $\hat \mu_i =  \frac{1}{N_i}\sum_{j=1}^{N_i}
\delta_{\x_{i,j}} $ with $\x_{i,1}, \x_{i,2}, \cdots \x_{i,N_i} \stackrel{i.i.d}{\sim} \mu_i$.  For a sake of simplicity, we assume in the sequel that the
number of samples for all distributions are equal to $N$.  
Suppose that we consider a dissimilarity $\mD$ and
that there exist a function $g_1$ such that $\mD$ satisfies a property of the form $\mathbf{P}\Big(\mD(\mu,\hat \mu)>\epsilon\Big)\leq 
g_1(K,N,\epsilon,d)$ then following theorem holds:

\begin{theorem} \label{th:main2} For a given learning problem, if the
  dissimilarity  $\mD$ is an $(\epsilon,\gamma)$-good
  dissimilarity function on population distributions, with
  $w(\mu) =1,\,\,\forall \mu$ and $K$ a parameter depending on this
  dissimilarity then, for a parameter $\lambda \in (0,1)$, if one
  draws a set $S$ from $\mathbb{P}$ containing
  $ n = \frac{32M^2}{\gamma^2} \log(\frac{2}{\delta^2 (1 - \lambda)})$
  positive examples $S^+ = \{\nu_1, \cdots, \nu_n\}$ and $ n$ negative
  examples $S^- = \{\zeta_1, \cdots, \zeta_n\}$, and from each
  distribution $\nu_i$ or $\zeta_i$, one draws
$N$ samples so that $\delta^2 \lambda \geq N g_1(K,N,\frac{\epsilon}{4},d)$
  samples so as to build empirical distributions $\{\hat \nu_i\}$ or
  $\{\hat \zeta_i\}$, then with probability $1 - \delta$, the mapping
  $ \hat \rho_S : \mathbb{P} \mapsto \R^{2n}$ defined as
  {
    $$\hat \rho_S(\hat \mu) = \frac{1}{M} (\mD( \hat \mu, \hat \nu_1 ),
    \cdots, \mD(\hat \mu, \hat \nu_n), \mD(\hat \mu, \hat \zeta_1 ),
    \cdots, \mD(\hat \mu, \hat \zeta_n ) )$$}
has the property that the induced distribution $\rho_S(\mathbb{P})$
  in $\R^{2n}$ has a separator of error at most $\epsilon + \delta$
  and margin at least $\gamma/4$.
\end{theorem}

Let us point out some relevant insights from this
theorem. At first, due to the use of empirical distributions
instead of population one, the
sample complexity of the learning problem increases for achieving
similar error as in Theorem \ref{th:main}. Secondly, note that $\lambda$ has a trade-off role on
the number $n$ of samples $\nu_i$ and $\zeta_i$ and the number of
observations per distribution. Hence, for a fixed error
$\epsilon + \delta$ at margin $\gamma/4$, having less samples per
distribution has to be paid by sampling more observations.

The proof of Theorem \ref{th:main} has been postponed to the appendix. It takes advantage of the following key technical result on empirical distributions.
\begin{lemma} \label{lem:bound}
 Let  $\mD$  be a dissimilarity on $\mathbb{P} \times \mathbb{P}$
such that $\mD$ is bounded by a constant $M$. 
Given a distribution $\mu \in \mathbb{P}$ of class $y$
and a set of independent distributions $\left\{\nu_{j}\right\}_{j=1}^n$, randomly drawn from
$\mathbb{P}$, which have
the same label $y$ and denote as $\hat \mu$ and $\{\hat \nu_i\}$ their empirical version composed of $N$ observations.  
Let us assume that there exists a function $g_1$ and 
a constant $K>0$ so that for 
any $\mu \in \mathbb{P}$,
$
\mathbf{P}\Big(\mD(\mu,\hat \mu)>\epsilon\Big)\leq 
g_1(K,N,\epsilon,d) 
$, with typically $g_1$ tends towards $0$ as $N$ or
$\epsilon$ goes to $\infty$. The following concentration inequality holds for any 
$\epsilon > 0$ :
\begin{align}\nonumber \tiny
   \mathbf{P}&\left ( \left|\frac{1}{n} \sum_{i=1}^n \mD(\hat \mu, \hat \nu_i)
- \E_{\nu \sim \mathbb{P}}[\mD(\mu,  \nu) | \ell(\mu) = \ell(\nu) ]
\right| > \epsilon \right) ~\\\nonumber
&\leq  N g_1(K,N, \frac{\epsilon}{4},d) +  2e^{-n \frac{\epsilon^2}{2M^2}}.
\end{align}
\end{lemma}
{This lemma tells us that, with high probability, the mean average 
of the dissimilarity between an empirical 
distribution and some other empirical distribution
{\em of the same class} does not differ much from the
expectation of this dissimilarity measured on population 
distributions.} 
Interestingly, the bound on the probability is composed
of two terms : the first one is related to the dissimilarity
function between a distribution and its empirical version 
while the second one is due to the empirical version of
the expectation (resulting thus from Hoeffding inequality).
The detailed proof of this result is given in the supplementary 
material. Note that in order for the bound to be informative,
we expect $g_1$ to have a negative exponential form  
in $N$. Another  version of this lemma is proven in supplementary where the concentration inequality for
the dissimilarity is on $|\mD(\hat \mu, \hat \nu) - \mD(\mu, \nu)|$.

From  a theoretical point of view, 
there is only one reason for choosing one $(\epsilon, \gamma)$-good dissimilarity function on population distributions from another. The rationale 
would be to consider the dissimilarity function with the fastest
rate of convergence of the concentration inequality $\text{Pr}(\mD(\mu, \hat
\mu) > \epsilon)$, as this rate will impact the upper bound in Theorem
\ref{th:main}.

From a more practical point of view, several factors may motivate
the choice of a dissimilarity function~: computational complexity
of computing $\mD(\hat \mu_i, \hat \mu_j)$, its empirical performance
on a learning problem and  adaptivity to different learning
problems ( \emph{e.g} without the need for carefully adapting its
parameters to a new problem.)

\section{($\epsilon-\gamma$) Good dissimilarity for distributions}

We are interested now in 
characterizing convergence properties of some dissimilarities (or
distance or divergence) on probability distributions
so as to make them fit into the framework.
Mostly, we will focus our attention on divergences
that can be computed in a non-parametric way.

\subsection{Optimal transport  distances}

Based on the theory of optimal transport, these distances
offer means to compare data probability distributions. More formally, 
assume that $\mX$ is endowed with a metric $d_\mX$. Let
$p\in(0,\infty)$, and let $\mu \in \mathbb{P}$ and
$\nu \in \mathbb{P}$ be two distributions with finite moments of order
p ({\em i.e.} $\int_\mX d_\mX(x,x_0)^pd\mu(x) < \infty$ for all $x_0$
in $\mX$). then, the p-Wasserstein distance is defined as:
 \begin{equation}
 W_p(\mu,\nu) = \left(\inf_{\pi \in \Pi(\mu,\nu)} \iint_{\mX\times \mX} d_\mX(x,y)^pd\pi(x,y)\right)^{\frac{1}{p}}.
 \label{eq:Wasserstein}
 \end{equation}
 Here, $\Pi(\mu,\nu)$ is the set of probabilistic couplings $\pi$ on
 $(\mu,\nu)$. As such, for every Borel subsets $A \subseteq \mX$, we
 have that $\mu(A)=\pi(\mX\times A)$ and $\nu(A)=\pi(A\times \mX)$. We
 refer to~\citep[Chaper 6]{Villani09} for a complete and
 mathematically rigorous introduction on the topic. 
Note when $p=1$, the resulting distance belongs to the
family of integral probability metrics \cite{sriperumbudur2010non}.
 OT has found numerous applications in machine learning domain such as multi-label classification
 \citep{Frogner15}, domain adaptation~\citep{courty2017} or generative
 models~\citep{arjovsky17a}.  Its efficiency comes from two major
 factors: {\em i)} it handles empirical data distributions without
 resorting first to parametric representations of the distributions
 {\em ii)} the geometry of the underlying space is leveraged through
 the embedding of the metric $d_\mX$. 
In some very  specific cases the solution of the infimum problem is analytic. For  instance, in the case of two Gaussians
 $\mu \sim \mathcal{N}(\m_1, \Sigma_1)$ and
 $\nu \sim \mathcal{N}(\m_2, \Sigma_2)$ the Wasserstein distance with
 $d_\mX(x,y)=\|x -y\|_2$ reduces to:
 \begin{equation}
 W^2_2(\mu,\nu)=||\m_1-\m_2||_2^2 + \mathcal{B}(\Sigma_1,\Sigma_2)^2
 \label{eq:W_Gauss}
 \end{equation}
 where  $\mathbb{B}(,)$ is the so-called Bures metric~\cite{bures1969extension}:
 \begin{equation}
 \mathcal{B}(\Sigma_1,\Sigma_2)^2=\text{trace}(\Sigma_1+\Sigma_2 - 2(\Sigma_1^{1/2}\Sigma_2\Sigma_1^{1/2})^{1/2}).
 \end{equation}
{If we make no assumption on the form of the distributions, and distributions are observed through samples, the Wasserstein distance is estimated by solving
a discrete version of Equation \ref{eq:Wasserstein} which is
a linear programming problem. }

One of the necessary condition for this distance to be relevant in our setting
is based on non-asymptotic deviation bound of the empirical distribution to the reference one. For our interest, \citet{fournier2015rate} 
have shown that for distributions with finite moments, the following
concentration inequality holds
$$\mathbf{P}(W_p(\mu,\hat \mu) > \epsilon) \leq C
\exp{(- K N \epsilon^{d/p})}$$
where $C$ and $K$ are constants that can be computed
from moments of $\mu$. This bound shows that the Wasserstein
distance suffers the dimensionality and as such a Wasserstein distance
embedding for distribution learning is not expected to be efficient
especially in high-dimension problems. However, a recent work
of \citet{weed2017sharp}  has also proved that under some hypothesis
related to singularity of $\mu$ better convergence rate
can be obtained (some being independent of $d$).
Interestingly, we demonstrate in what follows that 
the estimated Wasserstein distance for Gaussians using Bures metric
and plugin estimate of 
$\m$ and $\Sigma$ has a better bound related to the dimension.

\begin{lemma}
 Let  $\mu$ be a $d$-dimensional Gaussian distribution and
$\hat \m$ and $\hat \Sigma$ the sample mean and
covariance estimator of $\mu$ obtained from $N$ samples.
Assume that the true covariance matrix of $\mu$ satisfies
$\|\Sigma\|_2 \leq C_\Sigma$the random vectors $\mathbf{v}$ used for computing these estimates are so that $\|\v\|^2 \leq C_\mathbf{v}$ and
 The 
squared-Wasserstein distance between the empirical and
true distribution satisfies the following deviation
inequality:
\begin{align}\nonumber
\small
\mathbf{P}(W_2(\mu,\hat \mu)) > \epsilon) &\leq
2d \exp\Bigg(-\frac{-N \epsilon^2/(8d^4)}{C_\mathbf{v} C_\Sigma + 2C_\mathbf{v} \epsilon/(3d^2)} \Bigg)
\\\nonumber 
&~~+ \exp\Bigg(- \Big(\frac{N^{1/2}}{24\sqrt{C_\mathbf{v}}}\epsilon^2 - 1\Big)^{1/2} \Bigg) \nonumber
\end{align}
\end{lemma}
This novel deviation bound for the Gaussian 2-Wasserstein metric tells us
that if the empirical data (approximately) follows a high-dimensional Gaussian distribution then it  makes more sense to estimate the mean
and covariance of the distribution and  then to apply Bures-Wasserstein distance rather 
to apply directly a Wasserstein distance estimation based on the samples.

\subsection{Kullback-Leibler divergence and MMD}

Two of the most studied and analyzed divergences/distances 
on probability distribution are the Kullback-Leibler divergence
and Maximum Mean Discrepancy. Several works have proposed non-parametric approaches for estimating these distances and have provided theoretical
convergence analyses of these estimators.
 
For instance, \citet{nguyen2010estimating}  estimate the KL divergence
between two distributions by solving a quadratic programming problem
which aims a finding a specific function in a RKHS. They also proved
that the convergence rate of such estimator in in 
$\mathcal{O}(N^\frac{1}{2})$. MMD has originally been introduced
by \cite{gretton2007kernel} as a mean for comparing two distributions
based on a kernel embedding technique. It has been proved to be
easily computed in a RKHS. In addition, its empirical 
version benefits from nice uniform bound. Indeed, 
given two distribution $\mu$ and $\nu$ and their
empirical version based on $N$ samples $\hat \mu$ and $\hat \nu$,
the following inequality holds \citep{gretton2012kernel}:
$$
\mathbf{P}\Big( \text{MMD}^2(\hat \mu, \hat \nu) - \text{MMD}^2(\mu,\nu) > \epsilon \Big) \leq 
\exp{ \Big(-\frac{\epsilon^2 N_2 }{8K^2}\Big)}
$$ 
where $N_2 = \lfloor N/2 \rfloor$, $K$ is a bound on $k(\x,\x^\prime), \forall
\x, \x^\prime$, and $k(\cdot,\cdot)$ is the reproducing kernel of the RKHS
in which distributions have been embedded. We can note that this bound 
is independent of the underlying dimension of the data.

Owing to this property we can expect MMD to provide better estimation
of distribution distance for high-dimension problems than WD for
instance. Note however that for MMD-based two sample test,
\citet{ramdas2015decreasing} has provided contrary empirical evidence
and have shown that for Gaussian distributions, as dimension increases
$\text{MMD}^2(\mu,\nu)$ goes to $0$ exponentially fast in $d$. Hence,
we will postpone our conclusion on the advantage of one measure
distance on another to our experimental analysis.

\subsection{Discriminating normal distributions with the mean}
\label{sec:mean}

$(\epsilon, \gamma)$-goodness of a dissimilarity function is a
property that depends on the learning problem. As such, it is
difficult to characterize whether a dissimilarity will be good for all
problems. In the sequel we characterize this property for these three
dissimilarities, on  a mean-separated Gaussian distribution
problem.  \citep{muandet2012learning} used the same problem as their
numerical toy problem.
We show that even in this simple case,
MMD suffers high dimensionality more than the two other dissimilarities.

Consider a binary distribution classification problem where
samples from both classes are defined by Gaussian distributions in $\R^d$. 
 Means of these Gaussian distribution follow another Gaussian distribution
which mean depends on the class while covariance are fixed. Hence,
we have $\mu_i \sim \mathcal{N}(\m_i, \Sigma)$ with $\m_i \sim 
 \mathcal{N}(\m_{-1}^\star, \Sigma_0)$ if $y_i =  -1$ and $\m_i \sim 
 \mathcal{N}(\m_{+1}^\star, \Sigma_0)$ if $y_i =  +1$ where
$\Sigma$ and $\Sigma_0$ are some definite-positive covariance matrix. We suppose that both classes have same priors. We also denote $ D^\star = \|{\m_{-1}^\star} - \m_{+1}^\star\|_2^2$ which is a key component in the learnability 
of the problem. Intuitively, assuming that the
volume of each $\mu_i$ as defined by the determinant
of $\Sigma$ is smaller than the volume of $\Sigma_0$, the
larger $D^\star$ is the easier the problem should be. This
idea appears formally in what follows.

Based on Wasserstein distance between two normal distributions {with same covariance matrix}, we have
$W(\mu_i,\mu_j)^2 = \|\m_i - \m_j\|_2^2$. 
In addition, given a $\mu_i$ with mean $\m_i$, regardless of its
class, we have, with $k \in \{-1,+1\}$:
$$
\E_{\mu_j: \m_j \sim  \mathcal{N}(\m_{k}^\star, \Sigma_0)}[\|\m_i - \m_j\|_2^2] 
= \|\m_i - \m_k^\star\|_2^2 + \Tr(\Sigma_0)
$$
Given $\alpha \in ]0,1]$, we define the subset of $\R^d$, 
$$\mE_{-1} = \{\m :
(\m - \m_{-1})^\top (\m_{+1}^\star - \m_{-1}^\star) \leq \frac{1- \alpha}{2}  
D^\star$$
Informally, $\mE_{-1}$ is an half-space containing  of $\m_{-1}$ for
which all points are nearer to $\m_{-1}$ than $\m_{+1}$ with a margin
defined by  $\frac{1- \alpha}{2}  
D^\star$.  In the same way, we define $\mE_{+1}$ as :
$$\mE_{+1} = \{\m :
(\m - \m_{+1}^\star)^\top (\m_{-1}^\star - \m_{+1}^\star) \leq \frac{1- \alpha}{2}  
D^\star\}$$
Based on these definition, we can state that $W(\cdot,\cdot)$ is a $(\epsilon,\gamma)$ good dissimilarity
function with $\gamma = \alpha D^\star$,  $\epsilon = \frac{1}{2}\int_{\R^d \setminus \mE_{-1}}  d\mathcal{N}(\m_{-1}, \Sigma_0) + \frac{1}{2} \int_{\R^d \setminus \mE_{+1}}  d\mathcal{N}(\m_{+1}, \Sigma_0)$ and $w(\mu) = 1, \forall \mu$.
Indeed, it can be shown that for a given $\mu_i$ with $y_i = -1$, if
 $\m_i \in \mE_{-1}$ then 
$$
 \|\m_i - \m_{-1}^\star\|_2^2  +  \underbrace{\alpha \|\m_{-1}^\star - \m_{+1}^\star\|_2^2}_{\gamma} \leq  \|\m_i - \m_{+1}^\star\|_2^2 
$$
With a similar reasoning, we get an equivalent inequality for $\mu_i$ of
positive label. Hence, we have all the conditions given in Definition 
\ref{def:diss} for the Wasserstein distance to be an $(\epsilon,\gamma)$ good dissimilarity function for this problem. Note that the $\gamma$ and
$\epsilon$ naturally depend on the distance between expected means. 
The larger this distance is, the larger the margin and the smaller $\epsilon$
are.

Following the same steps, we can also prove that for this specific
problem of discriminating normal distribution, the Kullback-Leibler
divergence 
is also a $(\epsilon,\gamma)$ good dissimilarity function. Indeed, for $\mu_1$ and $\mu_2$ following two Normal distributions with same covariance matrix $\Sigma_0$, we have
$KL(\mu_1, \mu_2) = \|\m_2 - \m_1\|_{\Sigma_0^{-1}}^2$. And following
exactly the same steps as above, but replacing inner
product $\m^\top\m^\prime$ with $\m^\top\Sigma_0^{-1}\m^\prime$ leads
 to similar margin $\gamma = \alpha \|\m_{-1}^\star - \m_{+1}^\star\|_{\Sigma_0^{-1}}^2$
and similar definition of $\epsilon$.

While the above margins $\gamma$ for KL and WD are valid for 
any $\Sigma_0$, if we assume $\Sigma_0 = \sigma^2 \mathbf{I}$, 
then according to \cite{ramdas2015decreasing}, the following approximation
holds for this problem 
$$
\text{MMD}^2(\mu_1,\mu_2) \approx \frac{\|\m_1 - \m_2\|_2^2}{\sigma_k^2(1+2\sigma^2/\sigma_k^2)^{d/2+1}}
$$
where $\sigma_k$ is the bandwidth of the kernel embedding, leading to
a $(\epsilon-\gamma)$ distribution with margin $$\alpha \frac{\|\m_1^\star - \m_2^\star\|_2^2}{\sigma_k^2(1+2\sigma^2/\sigma_k^2)^{d/2+1}}
$$
From these margin equations for all the dissimilarities, we can drive
similar conclusions to those of \citet{ramdas2015decreasing} on test
power. Regardless on the choice of the kernel embedding bandwidth, the
margin of MMD is supposed to decrease with respect to the
dimensionality either polynomially or exponentially fast. As such,
even in this simple setting, MMD is theoretically expected 
to work worse than KL divergence or WD.

In practice, we need to compute these KL, WD or MMD distance from samples
obtained $\emph{i.i.d}$ from the unknown distribution $\mu$ and $\nu$.
The problem of estimating in a non-parametric way some
$\phi$-divergence, especially the Kullback-Leibler divergence have
been thoroughly studied by 
\citet{nguyen2007nonparametric,nguyen2010estimating}.  For KL
divergence, this estimation is obtained by solving a quadratic
programming problem.  
In a nutshell, compared to Kullback-Leibler divergence, Wasserstein distance
benefits from a linear programming problem compared to a quadratic programming
problem.
In addition, unlike KL-divergence, Wasserstein distance takes into account the
properties of $\mX$ and as such it does not diverge for distributions that
do not share support.

\section{Numerical experiments}

\begin{figure*}[t]
  \centering
\includegraphics[width=7cm]{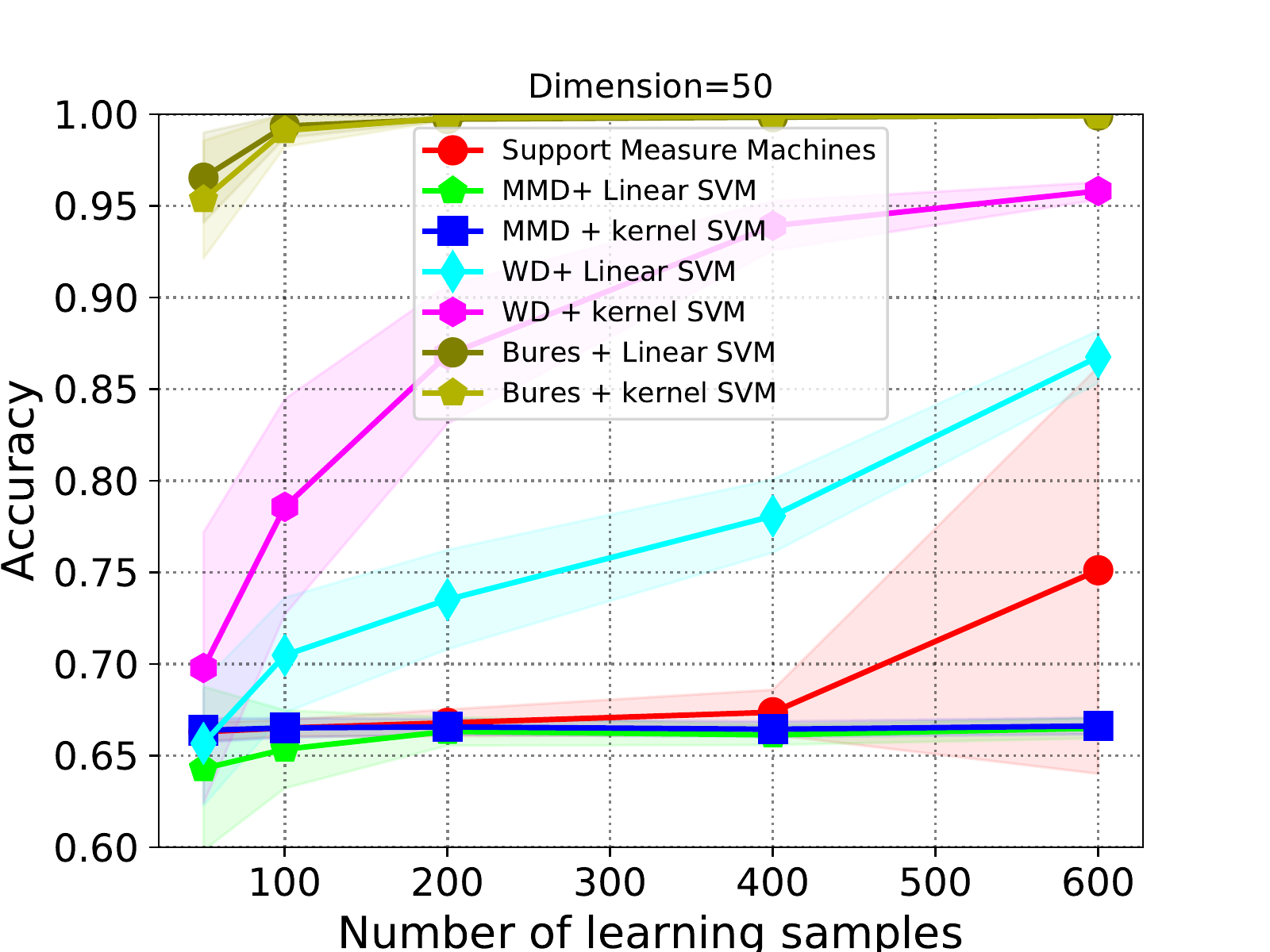}
\includegraphics[width=7cm]{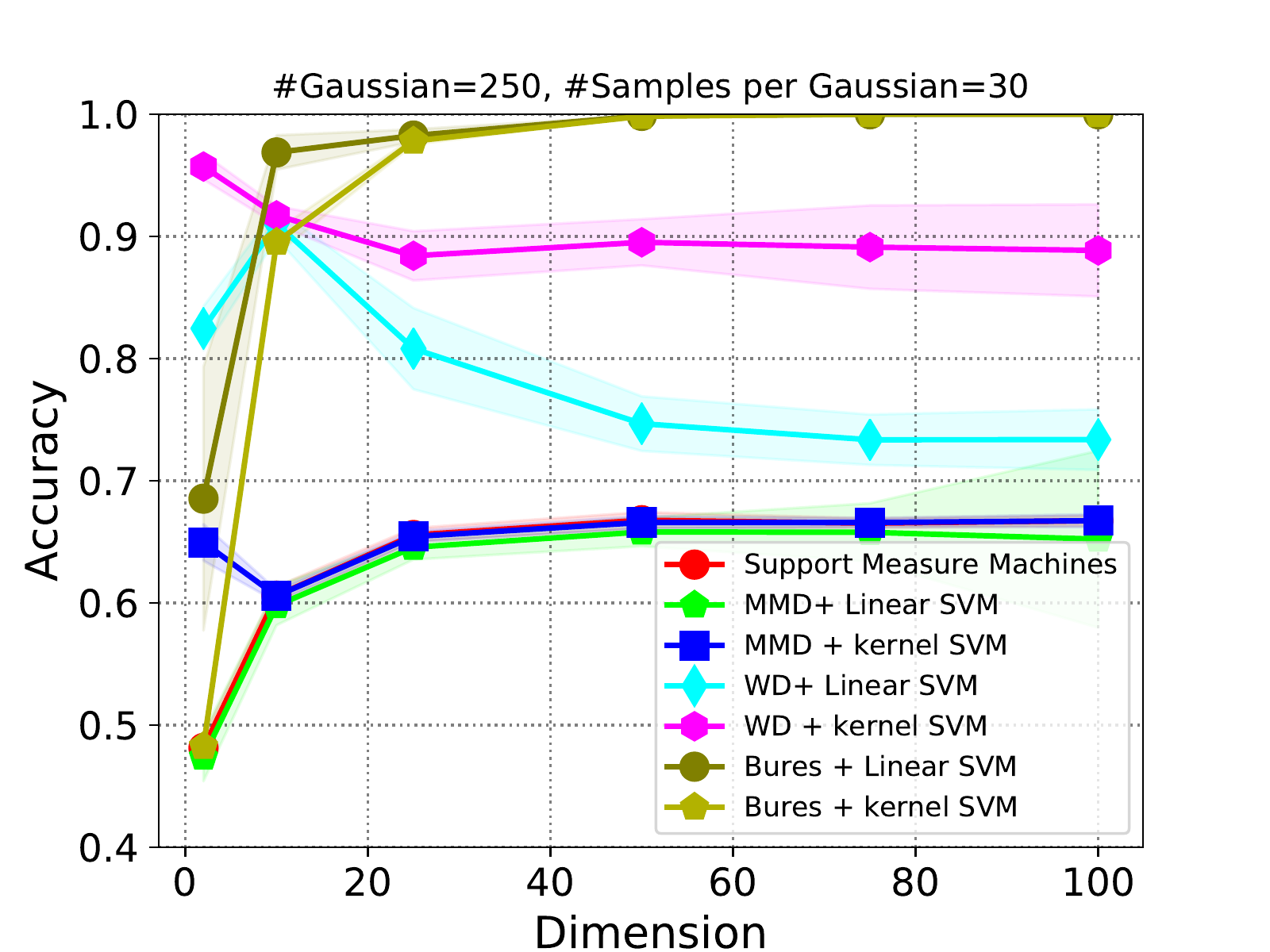}\\
  \caption{Comparing performances of Support Measure Machines and Wasserstein distance + classifier. 
}\label{fig:toy_res}
\end{figure*}

In this section, we have analyzed and compared the performances
of  Wasserstein distances based embedding for learning to
classify distributions. Several toy problems, similar
to those described in Section \ref{sec:mean} have been 
considered as well as a computer-vision real-world problem.

\subsection{Competitors}

Before describing the experiments we discuss the
algorithms we have compared. We have considered two variants of our
approach. The first one embeds the distributions based on 
$\hat \rho_S$ by using, unless specified, all  distributions
available in the training set. 
{The Wasserstein distance is approximated using its entropic
regularized version with $\lambda = 0.01$ for all problems.}
Then, we learn either a linear
SVM or a Gaussian kernel classifier resulting in two methods
dubbed in the sequel as \textbf{WD+linear } and
\textbf{WD+kernel}. {As discussed in section \ref{sec:empirical}, we can use the closed-form Bures Wassrestein distance when we suppose that the distributions are Gaussian. Assuming that
the samples come from a Normal distribution, plugging-in the empirical mean and
covariance estimation into the Bures-Wasserstein distance \ref{eq:W_Gauss}
gives us a distance that we can use as an embedding. In the
experiments, these approaches are named \textbf{Bures+linear}
and \textbf{Bures+kernel}.
}
In the family of integral probability metrics, we used the support measure machines of
\citet{muandet2012learning}, denoted
as \textbf{SMM}. We have considered its non-linear version
which used an Gaussian kernel on top of the MMD kernel. 
In SMM, we have thus two kernel hyperparameters.
{In order to evaluate the choice of the distance, 
we have also used the MMD distance in addition to the Wasserstein distance in our framework. These approaches are denoted
as \textbf{MMD+linear} and \textbf{MMD+kernel}.
}
Note that we have not reported based on samples-based approaches
such as SVM since \citet{muandet2012learning} have already
reported that they hardly handle distributions.

Kullback-Leibler divergence can replace the Wasserstein distance
in our framework. For instance, we have highlighted that for the 
 problem in Section \ref{sec:mean}, KL-divergence is an $(\epsilon,\gamma)$
 good dissimilarity function. We have thus implemented the
 non-parametric estimation of the KL-divergence based on quadratic programming
 \citep{nguyen2010estimating,nguyen2007nonparametric}.
After few experiments on the toy problems, we finally decided to not report  performance of the KL-divergence based approach
due to its  poor computational scalability as illustred in the supplementary
material.

\subsection{Simulated problem}

These problems aim at studying the performances of our models in
controlled setting. 
The toy problem corresponds to the one described in Section \ref{sec:mean}
but with $3$ classes. For all classes, mean of a given distribution
follows a normal distribution with mean $\m^\star =[1,1]$ and
covariance matrix $\sigma \mathbf{I}$ with $\sigma=5$. 
For class $i$, the covariance matrix of the distribution is
defined as $\sigma_i \mathbf{I} + u_i (\mathbf{I}_1 + \mathbf{I}_{-1}$)
where $\mathbf{I}_1$ and $\mathbf{I}_{-1}$ are respectively the
super and sub diagonal matrices. The $\{\sigma_i\}$ are constant
whereas $u_i$ follows a uniform distribution depending on the class. 
We have kept the number of empirical samples per distribution fixed at $N=30$.

For these experiments, we have analyzed the effect of the number of
training examples $n$ (which is also the number of templates) and the
dimensionality $d$ of the distribution.  Approaches are then evaluated on of $2000$  test distributions. $20$ trials have been considered for each $n$ and dimension $d$. We define a trial
as follows: we randomly sample the $n$ number of distributions and compute all
the embeddings and kernels. For learning, We have performed
cross-validation on all parameters of all competitors. This involves all kernel and classifier parameters. Details of all parameters and hyperparameters are given in the appendix.

Left plot in Figure \ref{fig:toy_res} represents the averaged classification accuracy with  $N=30$ samples per classes and $d = 50$ for increasing number
$n$ of empirical training distribution examples. Right plot
represents the same but for fixed $n=250$ and increasing dimensionality $d$.

From the left panel, we note that MMD-based distance fails in achieving
good performances regardless of how they are employed (kernel or
distance based classifier). WDMM performs better than MMD especially
as the number of training distrubtions increases. For $n=600$,
the difference in performance is almost $30\%$ of accuracy when
considering distance-based embeddings.
We also remark that the Bures-Wasserstein metric naturally fits to this Normal distribution learning problem and achieves perfect performances for $n \geq 200$.

Right panel shows the Impact of the dimensionality of the problem on the 
classification performance. We note that again MMD-based approaches 
do not perform as good as Wasserstein-based ones. Whereas MMD tops below $70\%$, our non-linear WDMM method achieves about $90\%$ of classification rate across a large range of dimensionality.

\subsection{Natural scene categorization}

\begin{table}[t]
  \centering
  \resizebox{\linewidth}{!}{\begin{tabular}{l|c|c|c}\hline 
  Method  & Scenes& 3DPC& 3DPC-CV\\ \hline\hline
SMM &  51.58 $\pm$ 2.46 & 92.79 &  92.99 $\pm$ 0.99  \\\hline 
MMD linear &  24.83 $\pm$ 1.22 & 91.89 &  91.84 $\pm$ 1.13 \\
MMD kernel &  27.02 $\pm$ 4.09 & 90.54 & 92.66 $\pm$ 1.02 \\\hline
WD linear &  \underline{61.58 $\pm$ 1.34} & \textbf{97.30} &\textbf{95.52 $\pm$ 0.89} \\
WD kernel & \underline{60.70 $\pm$ 2.49} &   96.86 & 94.89 $\pm$ 0.80 \\\hline
BW linear & \textbf{62.30 $\pm$ 1.32} &   64.86 &63.52 $\pm$ 5.72 \\
BW kernel & \underline{62.06 $\pm$ 1.34} &   70.72 & 72.20 $\pm$ 1.51 \\\hline \hline
  \end{tabular}}
\caption{Performances of competitors on real-world problems. 
3DPC and 3DPC-CV columns report performances on original train/test split and for random splits. Bold denotes best test accuracy
and underline show statistically equivalent performance under Wilocoxon
signrank test with $p=0.01$. 
}
\label{tab}
\vspace{-0.5cm}
\end{table}

We have also compared the performance the different approaches on
a computer vision problem. For this purpose, we have reproduced the 
experiments carried out by \citet{muandet2012learning}. 
Their idea is to consider an image of a scene as an histogram of
codewords, where the codewords have been obtained by k-means
clustering of $128$-dim SIFT vector and thus to use this histogram
as a discrete probability distribution for classifying the images.
Details of the feature extraction pipeline can be found in the paper
\citet{muandet2012learning}. The only difference our experimental
set-up is that we have used an enriched version of the dataset\footnote{The dataset is available at \url{http://www-cvr.ai.uiuc.edu/ponce_grp/data/}} they used. Similarly, we have used $100$ images per class for
training and the rest for testing. Again, all hyperparameters of all
competing methods have been selected by cross-validation. 

The averaged results over $10$ trials are presented in Table
\ref{tab}. Again, the plot illustrates the benefit of
Wasserstein-distance based approaches (through fully non-parametric
distance estimation or through the estimated Bures-Wasserstein metric)
compared to MMD based methods. We believe that the gain in performance
for non-parametric methods is due to the ability of the Wasserstein
distance to match samples of one distribution to only few samples of
the other distribution. By doing so, we believe that it is able to
capture in an elegant way complex interaction between samples of
distributions.

\subsection{3D point cloud classification}

3D point cloud can be considered as samples from a distribution.
As such, a natural tool for classifying them is to used  metrics or kernels
on distributions. In this experiment, we have benchmarked
all competitors on a subset of the ModelNet10
dataset \cite{}. Among the 10 classes in that dataset, we have
extracted the \emph{night stand}, \emph{desk} and \emph{bathtub} classes which respectively have $400$, $400$ and $212$ training examples and $172$,$172$ and
$100$ test examples. Experiments and model selection have been run
as in previous experiments. In Table \ref{tab}, we report results based on original train and test sets and results using $50-50$ random splits and resamplings. Again, we highlight the benefit of using the Wasserstein distance as an embedding and contrarily to other experiments, the Bures-Wasserstein metric yields
to poor performance as the object point clouds hardly fit a
Normal distribution leading thus to model misspecification.

\section{Conclusion}
This paper introduces a method for learning to discriminate probability
distributions based on dissimilarity functions. The algorithm
consists in embedding the distributions into a space of dissimilarity
to some template distributions and to learn a linear decision function
in that space. From a theoretical point of view, when considering
population distributions, our framework is an extension
of the one of \citet{balcan2008theory}. But we provide a theoretical analysis
showing that for embeddings based on empirical distributions, given enough
samples, we can still learn a linear decision functions with low error
with high-probability with empirical Wasserstein distance. The experimental results illustrate the benefits
of using empirical dissimilarity on distributions on toy problems
and real-world data.

Futur works will be oriented toward analyzing a more general class of regularized optimal transport divergence,
such as the  Sinkhorn divergence~\cite{Genevay17} in the context of Wasserstein distance measure machines. 
Also, we will consider extensions of this framework to regression problems, for which a direct 
application is not immediate.


\bibliographystyle{authordate1}

\begin{thebibliography}{}

\bibitem[\protect\citename{Arjovsky {\em et~al.\ }\relax, }2017]{arjovsky17a}
Arjovsky, Martin, Chintala, Sumit, \& Bottou, L{\'e}on. 2017.
\newblock {W}asserstein Generative Adversarial Networks.
\newblock {\em Pages  214--223 of:} {\em ICML}.

\bibitem[\protect\citename{Balcan {\em et~al.\ }\relax,
  }2008]{balcan2008theory}
Balcan, Maria-Florina, Blum, Avrim, \& Srebro, Nathan. 2008.
\newblock A theory of learning with similarity functions.
\newblock {\em Machine Learning}, {\bf 72}(1-2), 89--112.

\bibitem[\protect\citename{Bhattacharyya, }1943]{bhattacharyya1943measure}
Bhattacharyya, Anil. 1943.
\newblock On a measure of divergence between two statistical populations
  defined by their probability distributions.
\newblock {\em Bull. Calcutta Math. Soc.}, {\bf 35}, 99--109.

\bibitem[\protect\citename{Breiman, }2001]{breiman2001random}
Breiman, Leo. 2001.
\newblock Random forests.
\newblock {\em Machine learning}, {\bf 45}(1), 5--32.

\bibitem[\protect\citename{Bures, }1969]{bures1969extension}
Bures, Donald. 1969.
\newblock An extension of Kakutani's theorem on infinite product measures to
  the tensor product of semifinite $\sigma$-algebras.
\newblock {\em Transactions of the American Mathematical Society}, {\bf 135},
  199--212.

\bibitem[\protect\citename{Courty {\em et~al.\ }\relax, }2017]{courty2017}
Courty, Nicolas, Flamary, R{\'e}mi, Tuia, Devis, \& Rakotomamonjy, Alain. 2017.
\newblock Optimal transport for domain adaptation.
\newblock {\em IEEE Transactions on Pattern Analysis and Machine Intelligence}.

\bibitem[\protect\citename{Dietterich {\em et~al.\ }\relax,
  }1997]{dietterich1997solving}
Dietterich, Thomas~G, Lathrop, Richard~H, \& Lozano-P{\'e}rez, Tom{\'a}s. 1997.
\newblock Solving the multiple instance problem with axis-parallel rectangles.
\newblock {\em Artificial intelligence}, {\bf 89}(1-2), 31--71.

\bibitem[\protect\citename{Flaxman {\em et~al.\ }\relax,
  }2015]{flaxman2015supported}
Flaxman, Seth~R, Wang, Yu-Xiang, \& Smola, Alexander~J. 2015.
\newblock Who supported Obama in 2012?: Ecological inference through
  distribution regression.
\newblock {\em Pages  289--298 of:} {\em Proceedings of the 21th ACM SIGKDD
  International Conference on Knowledge Discovery and Data Mining}.
\newblock ACM.

\bibitem[\protect\citename{Fournier \& Guillin, }2015]{fournier2015rate}
Fournier, Nicolas, \& Guillin, Arnaud. 2015.
\newblock On the rate of convergence in Wasserstein distance of the empirical
  measure.
\newblock {\em Probability Theory and Related Fields}, {\bf 162}(3-4),
  707--738.

\bibitem[\protect\citename{Frogner {\em et~al.\ }\relax, }2015]{Frogner15}
Frogner, C., Zhang, C., Mobahi, H., Araya, M., \& Poggio, T. 2015.
\newblock Learning with a {W}asserstein Loss.
\newblock {\em In:} {\em NIPS}.

\bibitem[\protect\citename{{Genevay} {\em et~al.\ }\relax, }2017]{Genevay17}
{Genevay}, A., {Peyr{\'e}}, G., \& {Cuturi}, M. 2017.
\newblock {Learning Generative Models with Sinkhorn Divergences}.
\newblock {\em ArXiv e-prints}, June.

\bibitem[\protect\citename{Gretton {\em et~al.\ }\relax,
  }2007]{gretton2007kernel}
Gretton, Arthur, Borgwardt, Karsten~M, Rasch, Malte, Sch{\"o}lkopf, Bernhard,
  \& Smola, Alex~J. 2007.
\newblock A kernel method for the two-sample-problem.
\newblock {\em Pages  513--520 of:} {\em Advances in neural information
  processing systems}.

\bibitem[\protect\citename{Gretton {\em et~al.\ }\relax,
  }2012]{gretton2012kernel}
Gretton, Arthur, Borgwardt, Karsten~M, Rasch, Malte~J, Sch{\"o}lkopf, Bernhard,
  \& Smola, Alexander. 2012.
\newblock A kernel two-sample test.
\newblock {\em Journal of Machine Learning Research}, {\bf 13}(Mar), 723--773.

\bibitem[\protect\citename{Haasdonk \& Bahlmann, }2004]{haasdonk2004learning}
Haasdonk, Bernard, \& Bahlmann, Claus. 2004.
\newblock Learning with distance substitution kernels.
\newblock {\em Pages  220--227 of:} {\em Joint Pattern Recognition Symposium}.
\newblock Springer.

\bibitem[\protect\citename{Hein \& Bousquet, }2005]{hein2005hilbertian}
Hein, Matthias, \& Bousquet, Olivier. 2005.
\newblock Hilbertian metrics and positive definite kernels on probability
  measures.
\newblock {\em Pages  136--143 of:} {\em AISTATS}.

\bibitem[\protect\citename{Jebara {\em et~al.\ }\relax,
  }2004]{jebara2004probability}
Jebara, Tony, Kondor, Risi, \& Howard, Andrew. 2004.
\newblock Probability product kernels.
\newblock {\em Journal of Machine Learning Research}, {\bf 5}(Jul), 819--844.

\bibitem[\protect\citename{Muandet {\em et~al.\ }\relax,
  }2012]{muandet2012learning}
Muandet, Krikamol, Fukumizu, Kenji, Dinuzzo, Francesco, \& Sch{\"o}lkopf,
  Bernhard. 2012.
\newblock Learning from distributions via support measure machines.
\newblock {\em Pages  10--18 of:} {\em Advances in neural information
  processing systems}.

\bibitem[\protect\citename{Nguyen {\em et~al.\ }\relax,
  }2007]{nguyen2007nonparametric}
Nguyen, XuanLong, Wainwright, Martin~J, \& Jordan, Michael~I. 2007.
\newblock Nonparametric estimation of the likelihood ratio and divergence
  functionals.
\newblock {\em Pages  2016--2020 of:} {\em Information Theory, 2007. ISIT 2007.
  IEEE International Symposium on}.
\newblock IEEE.

\bibitem[\protect\citename{Nguyen {\em et~al.\ }\relax,
  }2010]{nguyen2010estimating}
Nguyen, XuanLong, Wainwright, Martin~J, \& Jordan, Michael~I. 2010.
\newblock Estimating divergence functionals and the likelihood ratio by convex
  risk minimization.
\newblock {\em IEEE Transactions on Information Theory}, {\bf 56}(11),
  5847--5861.

\bibitem[\protect\citename{Ntampaka {\em et~al.\ }\relax,
  }2015]{ntampaka2015machine}
Ntampaka, Michelle, Trac, Hy, Sutherland, Dougal~J, Battaglia, Nicholas,
  P{\'o}czos, Barnab{\'a}s, \& Schneider, Jeff. 2015.
\newblock A machine learning approach for dynamical mass measurements of galaxy
  clusters.
\newblock {\em The Astrophysical Journal}, {\bf 803}(2), 50.

\bibitem[\protect\citename{P{\'o}czos {\em et~al.\ }\relax,
  }2012]{poczos2012nonparametric}
P{\'o}czos, Barnab{\'a}s, Xiong, Liang, Sutherland, Dougal~J, \& Schneider,
  Jeff. 2012.
\newblock Nonparametric kernel estimators for image classification.
\newblock {\em Pages  2989--2996 of:} {\em Computer Vision and Pattern
  Recognition (CVPR), 2012 IEEE Conference on}.
\newblock IEEE.

\bibitem[\protect\citename{P{\'o}czos {\em et~al.\ }\relax,
  }2013]{poczos2013distribution}
P{\'o}czos, Barnab{\'a}s, Singh, Aarti, Rinaldo, Alessandro, \& Wasserman,
  Larry~A. 2013.
\newblock Distribution-Free Distribution Regression.
\newblock {\em Pages  507--515 of:} {\em AISTATS}.

\bibitem[\protect\citename{Ramdas {\em et~al.\ }\relax,
  }2015]{ramdas2015decreasing}
Ramdas, Aaditya, Reddi, Sashank~Jakkam, P{\'o}czos, Barnab{\'a}s, Singh, Aarti,
  \& Wasserman, Larry~A. 2015.
\newblock On the decreasing power of kernel and distance based nonparametric
  hypothesis tests in high dimensions.
\newblock {\em Pages  3571--3577 of:} {\em AAAI}.

\bibitem[\protect\citename{Sch{\"o}lkopf \& Smola,
  }2002]{scholkopf2002learning}
Sch{\"o}lkopf, Bernhard, \& Smola, Alexander~J. 2002.
\newblock {\em Learning with kernels: support vector machines, regularization,
  optimization, and beyond}.
\newblock MIT press.

\bibitem[\protect\citename{Sriperumbudur {\em et~al.\ }\relax,
  }2010]{sriperumbudur2010non}
Sriperumbudur, Bharath~K, Fukumizu, Kenji, Gretton, Arthur, Sch{\"o}lkopf,
  Bernhard, \& Lanckriet, Gert~RG. 2010.
\newblock Non-parametric estimation of integral probability metrics.
\newblock {\em Pages  1428--1432 of:} {\em Information Theory Proceedings
  (ISIT), 2010 IEEE International Symposium on}.
\newblock IEEE.

\bibitem[\protect\citename{Sutherland {\em et~al.\ }\relax,
  }2012]{sutherland2012kernels}
Sutherland, Dougal~J, Xiong, Liang, P{\'o}czos, Barnab{\'a}s, \& Schneider,
  Jeff. 2012.
\newblock Kernels on sample sets via nonparametric divergence estimates.
\newblock {\em arXiv preprint arXiv:1202.0302}.

\bibitem[\protect\citename{{V}illani, }2009]{Villani09}
{V}illani, {C}. 2009.
\newblock {\em {O}ptimal transport: old and new}.
\newblock {G}rund. der mathematischen {W}issenschaften.
\newblock {S}pringer.

\bibitem[\protect\citename{Weed \& Bach, }2017]{weed2017sharp}
Weed, Jonathan, \& Bach, Francis. 2017.
\newblock Sharp asymptotic and finite-sample rates of convergence of empirical
  measures in Wasserstein distance.
\newblock {\em arXiv preprint arXiv:1707.00087}.

\end{thebibliography}

\newpage
\appendix

\end{document}